\documentclass[11pt,a4paper]{article}
\usepackage[hyperref]{emnlp2020}
\usepackage{times}
\usepackage{latexsym}

\usepackage{graphicx}
\usepackage{booktabs}
\usepackage{amsmath}

\usepackage{microtype}

\aclfinalcopy 
\def\aclpaperid{920} 

\newcommand\norm[1]{\left\lVert#1\right\rVert}

\title{Acrostic Poem Generation}

\author{Rajat Agarwal \\
  New York University \\
  \texttt{rajat.agarwal@nyu.edu}
\And
  Katharina Kann \\
  University of Colorado Boulder \\
  \texttt{katharina.kann@colorado.edu}}

\date{}

\begin{document}
\maketitle
\begin{abstract}
  We propose a new task in the area of computational creativity: acrostic poem generation in English.
  Acrostic poems are poems that contain a hidden message; typically, the first letter of each line spells out a word or short phrase.
  We define the task as a generation task with multiple constraints: given an input word, 
  1) the initial letters of each line should spell out the provided word, 2) the poem's semantics should also relate to it, and 3) the poem should conform to a rhyming scheme.
  We further provide a baseline model for the task, which
  consists of a conditional neural language model in combination with a neural rhyming model. Since no dedicated datasets for acrostic poem generation exist, we create training data for our task by first training a separate topic prediction model on a small set of topic-annotated poems and then predicting topics for additional poems.
  Our experiments show that the acrostic poems generated by our baseline are received well by humans and do not lose much quality due to the additional constraints. Last, we confirm that poems generated by our model are indeed closely related to the provided prompts, and that pretraining on Wikipedia can boost performance.
\end{abstract}

\section{Introduction}
Poetry, derived from the Greek word $poiesis$ ("making"), is the art of combining rhythmic and aesthetic properties of a language to convey a specific message. Its creation 
is a manifestation of creativity, and, as such, hard to automate. However, since the development of creative machines is a crucial step towards real artificial intelligence, automatic poem generation is an important task at the intersection
of computational creativity and natural language
generation, and earliest attempts date back several decades; see \newcite{goncalo-oliveira-2017-survey} for an overview.

\begin{figure}[t!]
  \centering
  \includegraphics[width=.6\columnwidth]{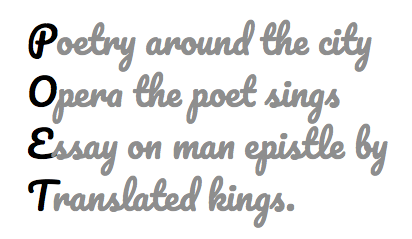}
  \caption{An acrostic poem generated by our proposed baseline model for the word \textit{poet}.}
  \label{fig:poem}
\end{figure}

With this paper, we add a new task to this research area: acrostic poem generation in English.
Acrostic poems, or simply \textit{acrostics}, are a special type of poetry,
in which typically the first letter of each line spells out a word or message, as in the example in Figure \ref{fig:poem}.
While this is the only formal characteristic of an acrostic, 
we here define the task of acrostic poem generation as generating poems such that, additionally, poems should also both rhyme and relate to the topic of their hidden word, e.g., the content of the poem in Figure \ref{fig:poem} should be related to the word "poet".
As far as meter is concerned, we are interested in free verse poems. 
Acrostic poem generation, as we define it, is a challenging \textit{constrained generation} task with multiple constraints: semantic ones (the content of the poem should follow a given topic), and structural ones (the poem should rhyme, and the first letters should spell out a given word). 

\begin{table*}[t!]
\centering
\setlength{\tabcolsep}{7.pt}
\begin{tabular}{l | rrrrr | r}
\toprule
\textbf{Number of lines} & \textbf{4} & \textbf{5} & \textbf{6} & \textbf{7} & \textbf{8} & \textbf{Total}\\
\midrule
\textbf{KnownTopicPoems} & 30,433 & 5,413 & 7,233 & 4,795 & 6,098 & 53,972\\
\midrule
\textbf{UnknownTopicPoems} & 26,986 & 10,765 & 11,609 & 6,433 & 9,487 & 65,280\\
\midrule
\textbf{Total} & 57,419 & 16,178 & 18,842 & 11,228 & 15,585 & \textbf{119,252}\\
\bottomrule
\end{tabular}
\caption{Number of poems in our datasets used for training, listed by the number of lines they contain.}
\label{tab:data_stats}
\end{table*}
We further propose a baseline model for the task, which we call the \textit{neural poet}. It is a generative model, which consists of two components: a conditional neural language model, which generates an acrostic poem based on a given word, and a rhyming model, trained on sonnets, which generates rhyming words for the last position in each line. 
Furthermore, two acrostic-specific challenges need to be solved: (i) generating such that the first letters of all lines spell out the defined word, and (ii) making sure that the resulting poem relates to the topic of that word.
We address the first challenge by limiting the choices in the output softmax during sampling from the language model. For the second challenge, we feed the word embedding of the topic to the language model at each time step. Since no large datasets for the task are available, we scrape poems for given key words from the web and train a separate discriminative model to predict topics, which we then use to predict silver standard topics for a larger poem dataset. Our final model is trained on a combination of poems with predicted and gold topics.   

Human evaluation on \textit{fluency}, \textit{meaningfulness}, and \textit{poeticness} \cite{manurung2004evolutionary} shows that our additional constraints hardly reduce performance as compared to unconstrained generation. Further, our poems relate to the acrostic word, even if the topic does not appear in the training set.
Finally, we show that model performance---in terms of
perplexity on a held-out validation set---can be improved by pretraining on Wikipedia.

\section{Datasets}
\label{sec:data}
To train the baseline model for our new task, we make use of 4 datasets, which we will describe here, before explaining the actual model in the next section.

\paragraph{KnownTopicPoems.}
We scrape poems from the web\footnote{\url{https://poemhunter.com/poem-topics}} in order to create our first dataset (KnownTopicPoems). The poems on this site are a good match for our task, since they are sorted by topics. Our resulting dataset contains 144 topics, and  a total of 32,786 poems. These poems are all of different lengths, but we aim at generating poems of up to 8 lines.\footnote{Extending our method to longer poems is straightforward.} Thus, we split all longer poems such that they contain at least 4 and a maximum of 8 lines in the following way: First, we split poems on empty lines, since those usually mark the end of a semantic unit. Second, if any of the resulting partial poems are still too long, we split them at full stops, but only use the beginning, since not every new sentence makes for a meaningful start of a poem. We add all possible options.  
This step increases the number of poems to 53,972, belonging to 144 topics. This dataset is used to train the language model of our neural poet, and to train the topic prediction model which is used to create additional training data for poem generation. We use 80\%, 10\%, and 10\% of the data for training, development, and test, respectively.

We tokenize all poems with the NLTK WordTreeBank tokenizer package \cite{Loper02nltk:the}. 

\paragraph{UnknownTopicPoems.}
We further make use of another poem dataset taken from \newcite{liu2018beyond}\footnote{\url{https://github.com/researchmm/img2poem/blob/master/data/unim_poem.json}} (UnknownTopicPoems), since it is larger than KnownTopicPoems. However, this dataset does not explicitly state the topics of individual poems. Thus, we automatically predict a topic for each poem with the help of our topic prediction model, which will be described in Subsection \ref{sub:topic}. The poems in UnknownTopicPoems are again broken down into poems with 4 to 8 lines, and, for our main experiments, this dataset is combined with KnownTopicPoems, increasing the number of samples to 119,252 poems for 144 topics. Table \ref{tab:data_stats} shows detailed statistics for both datasets. 
All poems are tokenized with the help of NLTK.

\paragraph{Sonnets.}
The last poem dataset we make use of is the sonnet poem dataset introduced by \newcite{lau2018deep}. This dataset consists of Shakespearean sonnets. Since those differ significantly in style from the poems in KnownTopicPoems and UnknownTopicPoems, we do not train our language model directly on them. However, we make use of the fact that sonnets follow a known rhyming scheme, and leverage them to train a neural model to produce rhymes, which will be explained in detail in Subsection \ref{sub:rhyme}. 
As for the previous two datasets, poems are tokenized using NLTK.

\paragraph{Wikipedia.}
Finally, we utilize a large English Wikipedia corpus\footnote{Version from 2018/10/01; downloaded from \url{https://linguatools.org/tools/corpora/wikipedia-monolingual-corpora/}} for pretraining. While this corpus does not consist of poems, we expect pretraining a language model on English text to help the overall coherence of the generated output.
Again, we tokenize all sentences using NLTK.

\section{The Neural Poet}
\begin{figure*}[t!]
  \centering
  \includegraphics[width=.85\textwidth]{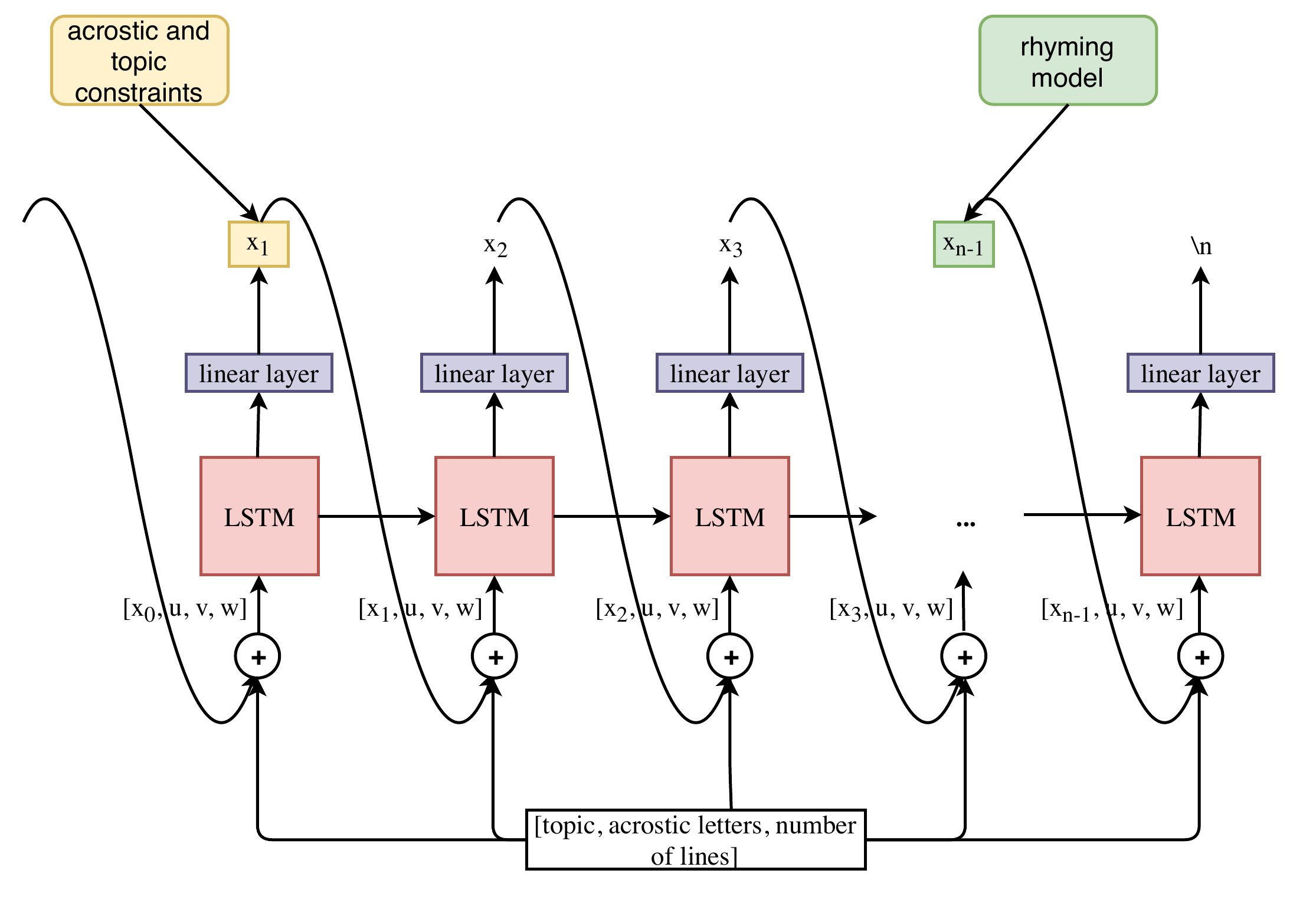}
  \caption{Overview of the baseline model we introduce together with the task of acrostic poem generation.}
  \label{fig:model}
\end{figure*}

We now describe all models that are either part of our baseline for acrostic poem generation or used for data preprocessing.
An overview of our final neural poet is shown in Figure \ref{fig:model}.

\subsection{Neural Poem Language Model}
\label{sub:lm}
We model the probability of a poem $x$, which consists of the word sequence $x_1x_2 ... x_n$, as:
\begin{align}
    p(x) &= \prod_{i=1}^n p(x_i | \{x_0, ... x_{i-1}\}, u, v, w)
\end{align}
$x_0$ is a start-of-sequence token, $u$ is a given topic, 
$v$ is the acrostic word, and $w$ is the number of lines the poem should consist of. 
We then model the conditional probability as
\begin{align}
    & p_{LM}(x_i | \{x_0, ... x_{i-1}\}, u, v, w) \nonumber\\
    = & g(\{x_0, ... x_{i-1}\}, u, v, w) \label{form:2}
\end{align}
for all but the first word in each line, since, for each first word, the probability depends on the acrostic word as described at the end of this subsection. 
The non-linear function $g$ is parameterized as a 3-layer uni-directional long short-term memory \citep[LSTM;][]{hochreiter1997long} language model. For generation -- but not during training --, $u$ and $v$ correspond to the same word.

At each time step, the model is given the last generated word, the topic of the poem, the characters of the acrostic word, and the number of lines to generate as input. 
Each letter of the acrostic word $v$ is represented as a one-hot vector of size 27.\footnote{We represent letters from $A-Z$, ignoring case, and add an additional padding token.} Our language model is trained to generate poems with up to 8 lines. Hence, the input matrix for the acrostic word is of size $8*27$, where, for poems corresponding to words with less than 8 letters, we make use of a padding token. All word-level input is represented by pretrained GloVe embeddings \cite{pennington2014GloVe}. The number of lines is represented in the model by a single-digit tensor.

\paragraph{The first word of each line.} 
The task of English acrostic poem generation as we define it demands that each line starts with a predefined letter. We want to enforce this constraint, while, at the same time, using the first word to guide the poem's content to stay close to the acrostic word. In order to achieve this with our baseline and still ensure coherent poem generation, we 
generate the initial words of each line as follows.

First, from all words in our vocabulary which start with the indicated character, we compute the $k=5$ nearest neighbors $n_1, \dots, n_k$ to the topic word $u$, using cosine similarity and our pretrained embeddings:
\[
\textrm{sim}(x, u) =\dfrac{\textrm{emb}(x) \cdot \textrm{emb}(u)}
{\norm{\textrm{emb}(x)} \cdot  \norm{\textrm{emb}(u)} }
\]

Then, we select our output with a probability of $m_1=0.7$ as 
\begin{align}
\textrm{argmax}_{i}\big(p_{LM}(n_1), p_{LM}(n_2), \dots, p_{LM}(n_k)\big)
\label{sim_score_equation}
\end{align}
However, this can cause the output to frequently become incoherent. Thus, we sample the first word from the language model, masking out all words that start with a wrong letter, with a probability of $m_2=0.3$.\footnote{When sampling from the nearest neighbors, we find it necessary to not permit end-of-line tokens as the second word in the line, since those are generated with a high frequency.}

\paragraph{The last symbol of each line.} Since the number of lines we want to generate is defined by the acrostic word, besides feeding an embedding which represents the number of lines to our model, we further enforce the right poem length by substituting each end-of-sentence symbol by end-of-line, if the number of lines is still too small. We do the same the other way around. Whenever this would lead to the last symbol of a poem being "," or ";", we substitute it by ".".

Thus, while our intuition is that having knowledge of the target number of lines will help our model for planning during generation, we do not require it to learn something that is already known.

\subsection{Rhyming Model}
\label{sub:rhyme}
We further make use of a separate neural model to generate rhyming words for the last position in each line. We aim at generating the following rhyming schemes. 4 lines: ABAB; 5 lines: ABABC; 6 lines: ABABCC; 7 lines: ABABCDC; 8 lines: ABABCDCD.

Whenever a rhyming word is required, our rhyming model computes the probability of an output word $c$, consisting of a character sequence $c_1c_2 ... c_l$, as:
\begin{align}
    p(c) &= \prod_{i=1}^l p(c_i | \{c_0, ... c_{i-1}\}, a, b) \nonumber\\
\end{align}
$a$ is the first word of the rhyme pair, i.e., the word $c$ should rhyme with, and $b$ denotes the poem up to $c$.
$a$ is represented as an encoding produced by concatenating the two last hidden states of a bidirectional character-level LSTM. $b$ is the last hidden state of a uni-directional character-level 
LSTM, which encodes the poem.

\paragraph{Training and hyperparameters. } Our rhyming model is trained on sonnet data (cf. Section \ref{sec:data}), since sonnets follow a known rhyming scheme. The LSTM which encodes the word to rhyme with has 1 hidden layer and 256-dimensional hidden states. The LSTM which encodes the poem also has 1 hidden layer, but 512-dimensional hidden states. 
Character embeddings are randomly initialized. The rhyming model is trained with Adam \cite{kingma2014adam} with an initial learning rate of $0.0005$, and a batch size of $64$.

\paragraph{Generation.} The words generated by the rhyming model substitute the last word in lines which are second in a rhyming pair. For instance, considering a 4-line poem which follows the scheme ABAB, the rhyming model would provide the last word of the third and fourth line, taking the last words of the first and, respectively, second line as input. During generation, we use beam search with width 5 to generate 5 candidates. Similarly to the line starts described above, we choose the candidate word with the highest language model probability as our final output.

\subsection{Topic Prediction Model}
\label{sub:topic}
Since our poem generator expects the topic of each poem, but the UnknownTopicPoems dataset does not provide any, we train a topic prediction model to create silver standard topic annotations.

We model the probability of poem $x$, consisting of the word sequence $x_1x_2 ... x_n$, belonging to topic $y$ as:
\begin{align}
    p(y) = d(\{x_0, x_1, ..., x_n, x_{n+1}\})
\end{align}
where $x_0$ is a start-of-sequence token, $x_{n+1}$ is an end-of-sequence token, and $d$ is a bidirectional word-level LSTM.

\paragraph{Training and hyperparameters. }
Our topic prediction LSTM has 1 hidden layer and a hidden size of 1024.
It is trained on the KnownTopicPoems dataset. We use Adam with an initial learning rate of $0.0005$  
and a batch size of $128$. 

\section{Experiments}
\subsection{Language Model Evaluation}
\begin{table}[t]
\centering
\setlength{\tabcolsep}{8.pt}
\begin{tabular}{l l}
\toprule
\textbf{Model} & \textbf{Perplexity} \\
\midrule
GOLD+ & 24.22\\
GOLD- & 23.79\\
PRED/GOLD+ & 19.94\\
PRED/GOLD- & 18.79\\
WIKI+ & \textbf{16.87}\\
WIKI- & 18.19\\
\bottomrule
\end{tabular}
\caption{Perplexity on the test set of KnownTopicPoems for all language models; best score in bold.}
\label{tab:ppl}
\end{table}
\paragraph{Experimental setup.}

\paragraph{Models.}
We train multiple language models to select the best basis for our neural poet:
\begin{itemize}
    \item \textbf{GOLD+ and GOLD-.} Our first networks are only trained on gold poems, i.e., our KnownTopicPoems dataset. "+" and "-" indicate if topics are fed into the model (+) or substituted by zero vectors (-).
    \item \textbf{PRED/GOLD+ and PRED/GOLD-.} We further train language models on both the KnownTopicPoems dataset and the UnknownTopicPoem dataset. "+" and "-" indicate if topics are fed into the model (+) or substituted by zero vectors (-).
    \item \textbf{WIKI+ and WIKI-.} Finally, we pretrain two language models on Wikipedia and finetune them on a combination of KnownTopicPoems and UnknownTopicPoem. Again, "+" and "-" indicate if topics are fed into the model (+) or substituted by zero vectors (-).
\end{itemize}

\begin{table*}[t!]
\centering
\setlength{\tabcolsep}{4.5pt}
\begin{tabular}{l | rrrr | rrrr | rrrr }
\toprule
& \multicolumn{4}{c}{\textbf{All}} & \multicolumn{4}{c}{\textbf{Known$^\heartsuit$}} & \multicolumn{4}{c}{\textbf{Unknown$^\spadesuit$}} \\
& \textbf{F} & \textbf{M} & \textbf{P} & \textbf{A} & \textbf{F} & \textbf{M} & \textbf{P} & \textbf{A} & \textbf{F} & \textbf{M} & \textbf{P} & \textbf{A} \\
\toprule
\textbf{Human} & 4.1 & 3.95 & 4.22 & 3.67 & 4.1 & 3.95 & 4.22 & 3.67 & - & - & - & - \\
\toprule
\textbf{NeuralPoet} & 3.48 & 2.75 & \textbf{3.66} & 2.55 & \textbf{3.70} & 2.86 & \textbf{3.77} & 2.79 & 3.25 & 2.63 & 3.56 & 2.31 \\
\midrule
\textbf{NeuralPoet-ST} & 3.51 & 2.79 & 3.25 & 2.59 & 3.39 & 2.81 & 3.31 & 2.73 & \textbf{3.62} & 2.76 & 3.20 & 2.43 \\
\midrule
\textbf{NeuralPoet-ST-AC} & \textbf{3.60} & 2.95 & 3.59 & 2.62 & 3.58 & \textbf{3.12} & 3.35 & 2.70 & \textbf{3.62} & 3.03 & \textbf{3.83} & 2.56 \\
\midrule
\textbf{NeuralPoet-ST-RH} & 3.36 & 2.94 & 3.32 & 2.54 & 3.40 & 2.99 & 3.41 & 2.69 & 3.32 & 2.89 & 3.27 & 2.38 \\
\midrule
\textbf{NeuralPoet-ST-TP} & \textbf{3.60} & \textbf{3.11} & 3.52 & \textbf{2.87} & \textbf{3.70} & 3.06 & 3.57 & \textbf{2.84} & 3.50 & \textbf{3.15} & 3.48 & \textbf{2.90} \\
\bottomrule
\end{tabular}
\caption{Human evaluation and ablation study;
\textit{F} = Fluency; \textit{M} = Meaning; \textit{P} = Poeticness; \textit{A} = Overall; \textit{ST}=selecting first words for each line according to the acrostic; \textit{AC}=acrostic forcing; \textit{RH}=rhyming model; \textit{TP}=feeding of topic vector.}
\label{tab:human_eval}
\end{table*}

\paragraph{Hyperparameters.}
All language models have 3 hidden layers and hidden states of size 1024 in all layers. Dropout \cite{srivastava2014dropout} of 0.4 is applied between layers for regularization. 100-dimensional GloVe embeddings are used to encode the input. The number of tokens in the GloVe embeddings are reduced
to the 50,000 most frequently occurring tokens in the dataset. 
We keep the GloVe and character embeddings fixed and do not update them during training.
All models are trained with early stopping with patience 25 on the development split from KnownTopicPoems.

\paragraph{Results.}
Results on the test split of KnownTopicPoems are shown in Table \ref{tab:ppl}
. WIKI+ and WIKI- obtain the lowest perplexity scores. Thus, we use them as the basis of our neural poet for the remaining experiments in this paper.

\begin{table}[th!]
\centering
\setlength{\tabcolsep}{17.pt}
\begin{tabular}{ll}
\toprule
\textbf{Known}$^\heartsuit$ & \textbf{Unknown}$^\spadesuit$ \\
\midrule
alone	&	bird	\\
fire	&	blizzard	\\
food	&	cake	\\
heaven	&	canyons	\\
hero	&	clever	\\
home	&	curse	\\
january	&	diary	\\
laughter	&	east	\\
loss	&	ending	\\
marriage	&	feather	\\
memory	&	general	\\
money	&	holiday	\\
music	&	local	\\
nature	&	song	\\
ocean	&	special	\\
respect	&	summer	\\
river	&	sweet	\\
star	&	tear	\\
thanks	&	tomorrow	\\
trust	&	width	\\
\bottomrule
\end{tabular}
\caption{The acrostic words used to generate poems in our experiments, corresponding to known or unknown topics.}
\label{tab:exp_topics}
\end{table}
\subsection{Human Evaluation Of Poems}
\paragraph{Experimental setup.}
In order to get an idea of the difficulty of our proposed task, we need to assess the quality of the poems generated by our baseline.
Following previous work \cite{manurung2004evolutionary,zhang-lapata-2014-chinese,loller2018deep},
we ask human annotators to evaluate 40 poems generated for the words in Table \ref{tab:exp_topics} for
\textit{readability} (lexical and syntactic coherence), \textit{meaningfulness}
(if the poem can be interpreted as conveying a message to its reader), and \textit{poeticness} (if the poem rhymes and looks like a poem) on a scale from 1 (worst) to 5 (best). Additionally, we also ask for an overall score. 
We collect  a minimum of 2 and a maximum of 5 ratings for each aspect of each poem. All annotators are fluent in English: they either are or have in the past been working or studying at an English-speaking institution.

\paragraph{Models.}
In order to further obtain insight into the effect of the different components of our model, we perform an ablation study: we evaluate our final neural poet including the rhyming model, a given topic, and acrostic forcing against versions of the model without selected components. 
We evaluate the following models:
\begin{itemize}
    \item \textbf{NeuralPoet.} This is our final model with all components. It has been pretrained on Wikipedia and fine-tuned on both KnownTopicPoems and UnknownTopicPoems.
    \item \textbf{NeuralPoet-ST.} This is NeuralPoet, but we do not choose the first word of each line from the nearest neighbors of the topic, i.e., we set $m_1=0$ and $m_2=1$ in the notation from Section \ref{sub:lm}. 
    \item \textbf{NeuralPoet-ST-AC.} Next, we additionally switch off acrostic forcing, i.e., we do not enforce that the first letters spell out the acrostic word. Since the language model receives embeddings of the characters as input, we observe that NeuralPoet-ST-AC still largely produces the acrostic word, but the language model has more freedom to generate coherent and fluent text.
    \item \textbf{NeuralPoet-ST-RH.} This is NeuralPoet-ST, but we switch off the rhyming model, i.e., the last words are generated directly from the language model. We expect this to also be more fluent and coherent, but less poem-like. 
    \item \textbf{NeuralPoet-ST-TP.} This version of  NeuralPoet-ST does not receive the topic as input. To achieve this, we set the topic vector to zero during both training and generation.
\end{itemize}
We further collect ratings for human poems from our training set for comparison; some of them are partial poems as used for training.

\begin{table}[t!]
\centering
\setlength{\tabcolsep}{7.pt}
\begin{tabular}{l}
\toprule
\vspace{.17cm}
\textbf{alone$^\heartsuit$} \\
\textbf{A}lone we spoke, \\
\textbf{L}ess, do not fear my heart, \\
\textbf{O}nly later, i may not love, \\
\textbf{N}ot to have hoped that i would not apart, \\
\textbf{E}ven... i am sure.\\
\midrule
\vspace{.17cm}
\textbf{nature$^\heartsuit$} \\
\textbf{N}ot still a child \\
\textbf{A}m i one of you \\
\textbf{T}hat look in the wild \\
\textbf{U}pon your paradise full of view \\
\textbf{R}emember my soul 's face well\\ 
\textbf{E}xperience 's as shall.\\
\midrule
\vspace{.17cm}
\textbf{cake$^\spadesuit$} \\
\textbf{C}hocolate wall and marble cup\\
\textbf{A}pples howl with golden hair\\
\textbf{K}itchen of the world they stir\\
\textbf{E}at bread and eat there.\\
\midrule
\vspace{.17cm}
\textbf{tear$^\spadesuit$} \\
\textbf{T}ear that out my soul, \\
\textbf{E}arth 's heart, \\
\textbf{A}ngry, up, \\
\textbf{R}ocks of death.\\
\midrule
\vspace{.17cm}
\textbf{ending$^\spadesuit$} \\
\textbf{E}verything is done at the random quality \\
\textbf{N}ext after the wide circuit of the past \\
\textbf{D}ays of the day \\
\textbf{I}n the end of the last \\
\textbf{N}ew york 's, complete words \\
\textbf{G}oing nowhere heard.\\
\bottomrule
\end{tabular}
\caption{Example poems generated by our model for the indicated topics and used in our evaluation.$^\spadesuit$ = unknown topic; $^\heartsuit$ = known topic.}
\label{tab:poem_examples}
\end{table}
\paragraph{Results.}
All ratings are displayed in Table \ref{tab:human_eval}.
We can see that human poems obtain the highest scores overall; they serve as a rough upper bound on the scores of our models.
We then compare the different versions of our neural poet 
on the basis of our four criteria.
We observe the following:
\begin{itemize}
    \item \textbf{NeuralPoet.}  This model performs well in most evaluations, and it has the highest performance among all models for fluency and poeticness for \textit{Known}. For \textit{All}, NeuralPoet obtains the best poeticness score. This shows the effectiveness of our proposed baseline.
    \item \textbf{NeuralPoet-ST.}  We  only notice small differences in the scores for poems generated by this model and the previous one.  
    However, differences in poeticness are relatively large: 0.41, 0.46, and 0.36 for \textit{All}, \textit{Known}, and \textit{Unknown}, respectively.
    We hypothesize that a reason for this might be that the generated words are not always closely related to the given topic. Thus, the poems loose context and cohesiveness more often, leading to a worse overall impression.
    \item \textbf{NeuralPoet-ST-AC.} Without enforcement of the acrostic constraints, the language model has more autonomy in selecting words. As expected, the results show that the poems generated are more fluent than those of most other models: it obtains the highest fluency scores for \textit{All} and \textit{Unknown}. However, differences to NeuralPoet are relatively small, showing that we do not lose much quality by enforcing acrostics.
    Another effect of fewer constraints is that the poems are more coherent: they get the highest meaningfulness scores for \textit{Known}, and the second highest for \textit{All}.
    \item \textbf{NeuralPoet-ST-RH.} This version of our model
    gets scores in the lower half for all individual evaluations. This clearly shows that rhyming seems to be evaluated highly by human annotators and that the rhyming model is an important component of our neural poet.
    \item \textbf{NeuralPoet-ST-TP.} This version of the model, which is not given a topic, obtains highest or close-to-highest scores for most individual evaluations. In particular, the generated poems seem to be more fluent and coherent than the alternatives. However, they do not relate to any specific topic, which probably causes the drop in quality for poeticness, where this model always performs worse than NeuralPoet.
\end{itemize}

\begin{table}[t]
\centering
\setlength{\tabcolsep}{3.5pt}
\begin{tabular}{l | r  r | r}
\toprule
& \textbf{Known$^\heartsuit$} & \textbf{Unknown$^\spadesuit$} & \textbf{Total} \\
\midrule 
\textbf{Correct} & 9 & 12 & 21\\ 
\midrule
\textbf{Disagreement} & 8 & 7 & 15 \\ 
\midrule
\textbf{Incorrect} & 3 & 1 & 4\\ 
\bottomrule
\end{tabular}
\caption{Number of poems correctly or incorrectly identified by human annotators as belonging to the given topic. \textit{Disagreement} denotes examples where the annotators selected different poems.}
\label{tab:topic_eval}
\end{table}
\subsection{Human Evaluation Of Topicality}
\paragraph{Experimental setup.}
Besides the first characters of all lines forming the input prompt, our acrostic poem generation task further requires that the content of the poem should relate to the input word.
In order to gain insight if we succeed with the latter, we further conduct the following evaluation: we create poems with (i) our final neural poet and (ii) a version of the poet that only sees zero vectors as topic vectors during training and generation. Both models are forced to generate acrostics. 
We then show poems generated by both models together with the acrostic word to human evaluators and ask which poem is more closely related to the topic.

\paragraph{Models.}
We compare poems generated by the following two models:
\begin{itemize}
    \item \textbf{NeuralPoet.} Our final baseline model with all components. It is pretrained on Wikipedia and fine-tuned on a combination of KnownTopicPoems and UnknownTopicPoems. 
    \item \textbf{NeuralPoet-ST-TP.} This version of our model, as described in the last subsection, has no component which takes the topic as input: the first word of the line is generated exclusively by the language model, and the topic vector is set to zero. It, thus, makes it possible to evaluate if the topic is indeed being reflected in our poems.
\end{itemize}

\paragraph{Results.} As shown in Table \ref{tab:topic_eval}, our annotators agree on the poem generated by NeuralPoet to be closer related to its topic for 21 out of 40 poems. In 15 cases, the two annotators disagree, and only in 4 cases they find the  poem generated by NeuralPoet-ST-TP, i.e., the model that does in fact not know about the topic, to be more similar to it. This indicates that our poems indeed confirm with the topic given by the acrostic word. 

Considering poems for known and unknown topics separately, we get a similar picture. Suprisingly, however, 12 out of 20 and 9 out of 20 poems are recognized correctly for know and unknown words, respectively. Thus, our model works well even for topics it has not seen during training. 

\section{Related Work}
Automated poetry generation has long been getting attention from researchers at the intersection of artificial intelligence and computational creativity. 
Even before the advent of deep learning, researchers used stochastic models and algorithms to generate poems \cite{queneau1961100,oulipo1981atlas,gervas2000wasp,manurung2004evolutionary}. 
With the advancements in deep learning, more and more researchers are exploring possibilities of training neural networks to generate poems which mimic human creativity. The authors of \newcite{lau2018deep} trained a model on generating Shakespearean sonnets. They used a hybrid word-character LSTM-based recurrent neural network to generate poems, and used separate rhythmic and rhyming models to enforce sonnet structure on the poems generated. All three component were trained in a multi-task fashion. Their crowd-work and expert evaluations suggested that the generated poems conformed to the sonnet structure, however lacked readability and coherent meaning. We make use of explicit representations of topics to address the poem coherence and readability concern: as our poems are generated  based on a topic, we expect them to be more coherent.\\
Authors of \newcite{wang2018generating} generated Chinese poems based on images, rather than topic words. They used a combination of a convolutional neural network (CNN) and a gated recurrent unit (GRU) to generate poems which related to the target image. They also generated acrostic poems, but used character-level modelling to achieve this -- which was simpler than in our case, since they worked with Chinese text where characters often correspond to entire words. Our preliminary experiments on English showed that character-level models learn easily to generate acrostics by themselves, however do not follow the topic as coherently as word-level models. 
\newcite{zhang-lapata-2014-chinese,zhang-etal-2017-flexible,yang2017generating,yi2018chinese,yichinese,yang2019generating} are other examples of work on generating Chinese poems, but did not focus on acrostics.

\newcite{ghazvininejad-etal-2016-generating,ghazvininejad-etal-2017-hafez} built a model to generate poems based on topics in a similar fashion to ours. However, they chose words related to a given topic to be the \textit{last} words in each line -- and to rhyme. For this, they built rhyming classes for an input topic first, from which rhyming pairs were chosen. The most obvious differences to our work are, however, that they produced poems following predefined stress patterns, while we are interested in free verse poems and that we generate acrostic poems, while they did not.  

Finally, some additional work on poem generation includes \newcite{yi-etal-2018-automatic}, who applied reinforcement learning to the problem, in order to overcome the mismatch of training loss and evaluation metric.

\section{Conclusion}
We introduce a 
new task in the area of computational creativity: acrostic poem generation in English. The task consists of creating poems with the following constraints:
1) the first letters of all lines should spell out a given word, 2) the poem's content should also be related to that word, and 3) the poem should conform to a rhyming scheme.
We further present a baseline for the task, based on a neural language model which has been pretrained on Wikipedia and fine-tuned on a combination of poems with gold standard and automatically predicted topics. A separate rhyming model is responsible for generating rhymes.

We perform a manual evaluation of the generated poems and find that, while human poets still outperform automatic approaches, poems written by our neural poet obtain good ratings. Our additional constraints only slightly decrease fluency and meaningfulness and, in fact, even increase the poeticness of the generated poems. Furthermore, our model's poems are indeed topic-wise closely related to the acrostic word. Our neural poet is  available at \url{https://nala-cub.github.io/resources} as a baseline for future research on the task.

\section*{Acknowledgments}
We would like to thank the members of NYU's ML$^2$ group for their help with the human evaluation and their feedback on our paper! We are also grateful to the anonymous reviewers for their insightful comments.

\bibliographystyle{acl_natbib}
\bibliography{anthology,emnlp2020}

\onecolumn

\section*{Appendix A: Details on Computing}
\label{sec:appA}

\begin{table}[h]
\centering
\setlength{\tabcolsep}{3.5pt}
\begin{tabular}{l | r  r  r r }
\toprule
Model & Training Time & Number Epochs & Number Parameters \\
\midrule
Wikipedia pretraining & 600 minutes & 3 & 2298764 \\
Sonnet pretraining & 100 minutes & 30 & 1833642 \\
Neural poet & 2440 minutes & 50 & 1833642 \\
Rhyming model & 60 minutes & 50  & 36788 \\
Topic prediction model & 60 minutes & 50 & 140244 \\
\bottomrule
\end{tabular}
\caption{Training times and number of parameters for our models. All models have been trained with a batch size of 128 on an NVIDIA Titan V GPU with 12 GB RAM.}
\label{tab:trainingdetails}
\end{table}

\section*{Appendix B: Details on Hyperparameters}
\label{sec:appB}
Our hyperparameters have been manually tuned over small sets of intuitive values. 

\end{document}